# Can Modern NLP Systems Reliably Annotate Chest Radiography Exams? A Pre-Purchase Evaluation and Comparative Study of Solutions from AWS, Google, Azure, John Snow Labs, and Open-Source Models on an Independent Pediatric Dataset

Shruti Hegde MS, Mabon Manoj Ninan BS, Jonathan R. Dillman MD, MSc, Shireen Hayatghaibi PhD, Lynn Babcock MD, Elanchezhian Somasundaram PhD

## Abstract

**Purpose:**

General purpose clinical natural language processing tools are increasingly used for the automatic labeling of clinical reports to support various clinical, research and quality improvement applications. However, independent performance evaluations for specific tasks, such as labeling pediatric chest radiograph reports, remain scarce. This study aims to compare four leading commercial clinical NLP systems for entity extraction and assertion detection of clinically relevant findings in pediatric chest radiograph reports. In addition, the study evaluates two dedicated chest radiograph report labelers, CheXpert and CheXbert, to provide a comprehensive performance comparison of the systems in extracting disease labels defined by CheXpert.

**Methods:**

A total of 95,008 pediatric chest radiograph (CXR) reports were obtained from a large academic pediatric hospital for this IRB-waived study. Clinically relevant terms were extracted using four general-purpose clinical NLP systems: Amazon Comprehend Medical (AWS), Google Healthcare NLP (GC), Azure Clinical NLP (AZ), and SparkNLP (SP) from John Snow Labs. After standardization, entities and their assertion statuses (positive, negative, uncertain) from the findings and impression sections were analyzed using descriptive statistics, paired t-tests, and Chi-square tests. Entities from the Impression sections were mapped to 12 disease categories plus a *No Findings* category using a regular expression algorithm. In parallel, CheXpert and CheXbert processed the same reports to extract the same 13 categories (12 disease categories and a *No Findings* category). Outputs from all six models were compared using Fleiss' Kappa across the assertion categories. Model-specific performance was evaluated by comparing each system's assertions to a consensus pseudo ground truth generated by majority voting, with assertion accuracy reported per disease label and aggregated across all labels.

**Results:**

Across 95,008 pediatric CXR reports, a significant difference in the mean number of entities extracted per report was observed between the four commercial NLP systems (p-value: <0.001). The number of unique disease related entities detected by each system also showed large differences, with SP extracting 49,688 unique entities, GC 16,477, AZ 31,543, and AWS 27,216. Further, Significant differences in the distribution of entities classified as positive, negative, or uncertain were observed between the NLP systems for both the findings and impression sections of the report (p < 0.001). For the CheXpert labels comparison, the mean Fleiss' Kappa, calculated across all assertion categories (positive, negative, uncertain, and absent) for CheXpert labels, was 0.68 ± 0.17 when including all exams and 0.35 ± 0.14 when excluding exams where all six models predicted the disease as absent. The mean assertion accuracy across all six models, when compared with the consensus ground truth, was 62 ± 9%, with SP achieving the highest accuracy at 76% and AWS the lowest at 50%. CheXpert and CheXbert both had an accuracy of 56%.

**Conclusions:**

Our findings reveal significant variability in entity extraction performance and uncertainty levels among the evaluated clinical NLP systems. Although each system exhibits unique strengths, the study highlights the need for further research and manual review when deploying NLP technology in clinical report analysis.

**Clinical Relevance:**

The results offer important insights for healthcare professionals and researchers seeking to leverage NLP technology to enhance decision support and advance clinical research.

**Introduction:**

Patient Electronic Health Records (EHRs) have become central to discovery and innovation in data-driven healthcare [1]. EHRs encompass structured data such as diagnostic codes, physiological measurements, and medication records, as well as unstructured free-text clinical health records (e.g., office visit, inpatient, and surgical notes and imaging reports), which provide valuable insights to enhance care provision and inform clinical decision-making [2]. Natural Language Processing (NLP) has emerged as a crucial tool for extracting insights from these unstructured clinical reports [3], a task that is highly tedious when performed manually. Advancements in NLP have facilitated various clinical and research applications, including automated clinical coding [4], clinical decision support systems [5], predictive analytics for patient outcomes [6, 7], and population health management [8]. Additionally, NLP enables the extraction of diagnostic information from radiology reports [9], enhances disease surveillance [10], supports large-scale clinical research [7], improves patient care through personalized treatment recommendations [11] and enables quality improvement efforts.

NLP systems for processing clinical text can be categorized into rule-based and statistical learning-based approaches. Rule-based systems utilize medical ontology libraries that host expert-curated knowledge bases encompassing medical concepts, diagnostic codes, and medication categories [12]. In contrast, machine learning (ML) and hybrid (ML + rule-based) tools have been increasingly adopted for general-purpose clinical NLP applications [13, 14]. Notable among these are recurrent neural networks (RNNs) and long short-term memory (LSTM) networks, which have significantly advanced the handling of sequential data [15]. However, the introduction of attention mechanisms through the Transformer architecture has surpassed previous benchmarks, establishing Transformers as the industry standard in NLP [16]. Transformer-based methods have led to the development of two primary model families: the GPT (Generative Pre-trained Transformer) family [17], excelling in applications such as question answering and summarization through autoregressive generation, and the BERT (Bidirectional Encoder Representations from Transformers) family [18], optimized for tasks like named entity recognition, assertion detection, entity linking, and relationship extraction. Named Entity Recognition (NER) is the task of identifying and classifying specific pieces of information (called entities) in unstructured text into predefined categories (e.g., disease, procedure entities). Assertion detection determines the status or context of a recognized entity in clinical text (e.g., positive/negative/uncertain). Relationship Extraction is the task of identifying and classifying semantic relationships between two or more entities mentioned in unstructured text. In the clinical context, this means detecting how entities are connected, such as recognizing that a medication is prescribed to treat a specific condition.

Chest radiography (CXR) is pivotal for diagnosing and monitoring respiratory and cardiovascular conditions such as pneumonia [19], tuberculosis [20], lung cancer [21], and heart failure [22]. CXR reports offer detailed diagnostic information that aids in identifying and tracking these conditions, thereby supporting informed treatment and patient management decisions. However, interpreting CXR images is inherently complex, with studies revealing significant variability among physicians and radiologists in their assessments [23, 24]. This complexity makes CXR reports an ideal use case for evaluating NLP systems, as they encapsulate nuanced diagnostic information that challenges automated extraction and analysis.

The availability of several large public CXR datasets with images, reports, and disease labels has spurred the development and assessment of advanced computer vision, NLP, and multimodal AI models tailored specifically to CXR applications [25, 26]. Specifically, the CheXpert framework [27], which includes a comprehensive dataset of CXR images, reports, and associated disease labels, provides multiple report labeling models that are used to generate labels for other public datasets, such as the MIMICS CXR dataset [28]. Numerous research studies rely on these datasets to develop and evaluate their model performance [29-31]. Concurrently, general-purpose clinical NLP systems, ranging from open-source rule-based and Machine Learning models like MetaMap [32] and cTAKES [33] to recent commercial systems employing proprietary algorithms, are widely utilized for various clinical note processing tasks [34]. However, the performance of these more general systems on specific note types and tasks, such as pediatric CXR report labeling, is not well documented. This lack of standardized, independent comparison leaves users unaware of inherent errors in these systems and how such inaccuracies may propagate into downstream applications. A rigorous, standardized comparison of competing NLP systems on independent datasets is crucial to understanding their limitations, assessing their uncertainties, and ensuring more reliable integration into clinical workflows.

The primary objective of this study is to compare four commercial clinical NLP systems : Amazon Comprehend Medical (AWS) [35], Google Healthcare NLP (GC) [36], Azure Clinical NLP (AZ) [37], and SparkNLP (SP) from John Snow Labs [38], for extracting clinically relevant entities and determining their assertion status from an independent database of pediatric CXR reports. The secondary objective is to evaluate the performance of benchmark CXR report labeling models, CheXpert [27] and CheXbert [39], commonly used in CXR research, against an extraction pipeline constructed using general-purpose clinical NLP systems on this independent pediatric dataset.

**Methods:**

Figure 1 shows the overall methodology of this analysis. Patient consent for this retrospective study was waived by the Institutional Review Board, which approved this study. All exams were de-identified of Protected Health Information (PHI) by the Radiology Informatics team.

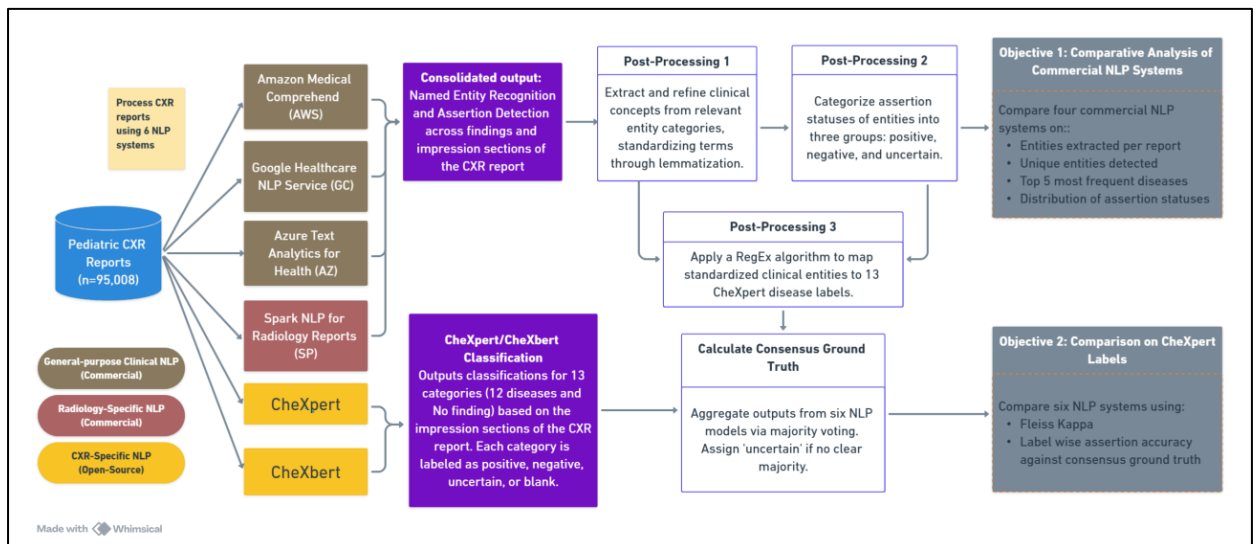

*Figure 1: This flowchart illustrates the methodology starting from pediatric CXR report extraction, followed by processing using the six NLP systems. Postprocessing steps include 1) entity standardization, 2) assertion categorization, and 3) aggregation into a consensus ground truth via majority voting. Performance of the models in entity extraction and assertion detection on all entities extracted by the commercial NLP systems and on CheXpert specific disease labels were analyzed.*

**a) Dataset:**

Pediatric CXR reports were extracted from the radiology information system at a large pediatric hospital for the study period spanning June 2015 to June 2020. A total of 95,008 examinations constituted the study sample for this study's evaluation. The mean age of the participants was 7.5 ± 7.5 years, with a male-to-female ratio of 1:2 (50503 males, 44286 females, 219 unknown). The CXR reports were typically in a semi-structured format, comprising the following sections: 1. Clinical History, 2. Findings, and 3. Impression. Each CXR radiology report was stored as an individual text file, linked via anonymized identifiers to ensure patient confidentiality.

**b) General purpose clinical NLP systems:**

This study compares four commercial clinical NLP systems: AWS [35], GC [36], AZ [37], and SP's NLP models for radiology reports [38]. The clinical NLP systems for AWS, AZ, and GC are accessed via their respective cloud-based Application Programming Interfaces (APIs), all of which are compliant with the Health Insurance Portability and Accountability Act (HIPAA, USA). Deidentified radiology reports were submitted to these APIs, and the output results were returned in JSON format. Each of the three cloud-based systems performed four major tasks: entity extraction, assertion detection, entity linking, and relationship extraction. [40, 41]

In contrast, the Spark NLP (SP) radiology models were accessed through an academic license and operated on local hardware. The extraction pipeline for SP consisted of BERT-based radiology models specifically designed for entity extraction and assertion detection. Unlike the cloud-based systems, the SP models are tailored to handle the unique nuances of radiology reports.

**c) Entity recognition:**

The output data structure and the number of entity categories extracted were unique to each NLP system. A custom postprocessing pipeline was developed to handle the JSON outputs from each system. The pipeline focused exclusively on disease-related entity categories and filtered out entities in other categories for the analysis. Table 1 provides the total number of named entity categories extracted by each system and the entity categories that were filtered for analysis. The categories for analysis are defined so that they are synonymous with disease symptoms or findings. The extracted entities from each NLP system were also normalized using a lemmatization algorithm such that related terms and phrases appear consistent for calculating descriptive statistics.

*Table 1: Details of the commercial NLP systems used in this study, the number of named entity categories detected by each system, and the selected categories that relate to disease symptoms or findings.*

| NLP system | Named Entity Categories | Categories Selected for Analysis |
|---|---|---|
| **AWS** – Amazon Medical Comprehend [v1.0.0, DetectEntities] | 7 | 'MEDICAL_CONDITION' |
| **AZ** – Azure Text Analytics for Health | 36 | 'SYMPTOM_OR_SIGN', 'DIAGNOSIS' |
| **GC** – Google Healthcare NLP system | 28 | 'PROBLEM' |
| **SP** – John Snow Labs Spark NLP [ner_radiology model] | 13 | 'Disease_Syndrome_Disorder', 'Symptom', 'ImagingFindings' |

### d) Standardization of assertion status:

Assertion refers to the context in which entities are mentioned in a clinical report, which is crucial for effective use of clinical notes in downstream applications. Each of the four commercial NLP systems identified assertions for extracted entities in different ways. Except for AWS, the other systems provided a categorical assertion status with unique definitions for each category. The AWS model provided only a numerical confidence score (CS) for the negation attribute associated with each detected entity. For a standardized comparative analysis, assertions for each entity were consolidated into three categories: positive, negative, or uncertain. Table 2 outlines the original assertion categories for each NLP system and how they were mapped into these three standardized assertion statuses for analysis.

*Table 2 Standardization criteria for each NLP system, mapping assertion outputs to three standardized assertion statuses: Positive, Negative, and Uncertain for analysis. CS is the negation confidence score for AWS.*

| NLP System | Positive | Negative | Uncertain |
|---|---|---|---|
| **AWS** – Amazon Medical Comprehend [v1.0.0, DetectEntities] | *CS < 0.25 | *CS > 0.75 | 0.25>= *CS <= 0.75 |
| **AZ** – Azure Text Analytics for Health | 'positive' | 'negative' | 'positivepossible', 'negativepossible', 'neutralpossible' |
| **GC** – Google Healthcare NLP system | 'likely' | 'unlikely' | 'somewhat_likely', 'somewhat_unlikely', 'uncertain', 'conditional' |
| **SP** – John Snow Labs Spark NLP [ner_radiology model] | 'present', 'none', 'past' | 'absent', 'family', 'someone else', 'planned' | 'hypothetical', 'possible' |

*CS stands for confidence score

### e) CheXpert disease labels and models:

This study utilized two open-source specialized models, CheXpert and CheXbert, to extract disease labels from the impression section of the chest radiograph (CXR) reports based on a predefined set of 13 categories relevant to chest radiography. These categories include *enlarged cardiomediastinum, cardiomegaly, lung opacity, lung lesion, edema, consolidation, pneumonia, atelectasis, pneumothorax, pleural effusion, pleural other, fracture,* and a *No Findings* category indicating exams without known disease findings. CheXpert is a rule-based model that employs heuristic rules and expert-curated mappings to assign disease labels from the textual content of CXR reports. In contrast, CheXbert is built on a BERT-based architecture that leverages deep contextual embeddings for a more nuanced interpretation of CXR reports, demonstrating improved accuracy on the CheXpert dataset. Both models produce labels accompanied by three assertion statuses positive, negative, and uncertain for each of the 13 output categories.

### d) Mapping clinical entities to CheXpert disease categories:

Lemmatization is a process in natural language processing (NLP) that reduces a word to its base or dictionary form, also known as the lemma (e.g., diagnosed, diagnoses and diagnosing would be reduced to base word diagnose). The extracted named entities from the commercial NLP systems, even after lemmatization, exhibited multiple variations for the same clinical concepts, depending on the system. To map these varied entities to the disease labels identified by the CheXpert models, a comprehensive regular expression (RegEx) algorithm was developed. This algorithm was primarily designed using the dictionary of keywords employed by the CheXpert model and was expanded by reviewing phrases, abbreviations, and linguistic patterns specific to the pediatric dataset that correspond to each disease category. By filtering and grouping relevant entities under the appropriate disease labels, the RegEx algorithm provided a consistent and unified categorization across the dataset.

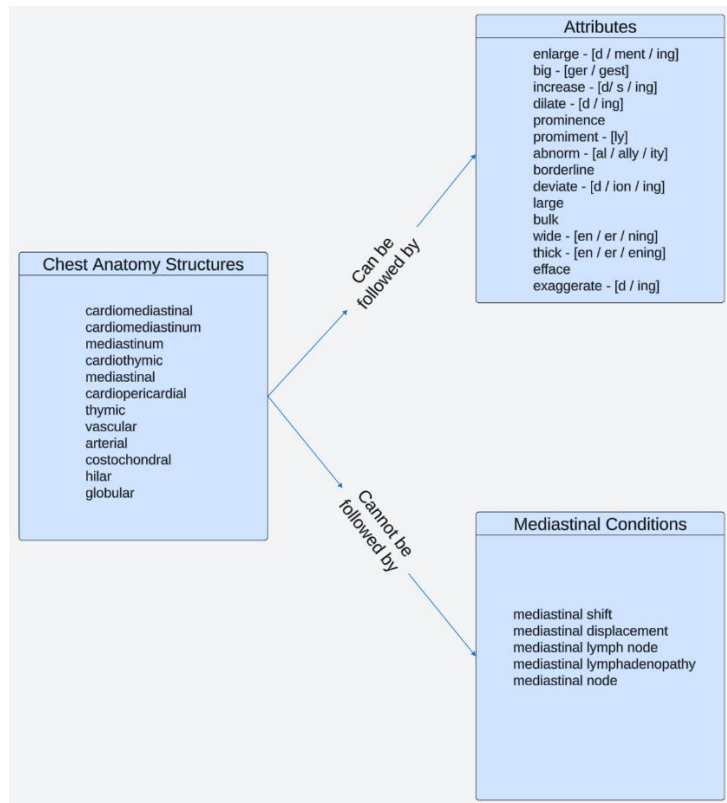

*Figure* 2: *This image illustrates the regex (regular expression) used to search for phrases and patterns related to the "enlarged cardiomediastinum" condition. The Chest Anatomy Structures list the base keywords that must be present, followed by Attributes that describe the anatomical structures. This combination forms a valid pattern for identifying "enlarged cardiomediastinum". However, if Chest Anatomy Structures are followed by certain Mediastinal Conditions, the pattern is considered invalid and is excluded from the "enlarged cardiomediastinum" search.*

Figure 2 illustrates how the algorithm detected *"enlarged cardiomediastinum",* a condition not often directly mentioned in CXR reports is labeled by identifying related phrases and patterns. The complete details of the RegEx patterns used for each disease label are provided in Appendix A. This standardization was critical for enabling reliable aggregation, comparison, and subsequent analysis of disease labels detected by the commercial NLP systems against the CheXpert and CheXbert models.

**e) Evaluation metrics:**

To evaluate the performance of commercial NLP systems in identifying disease-related entities from chest X-ray (CXR) reports, we analyzed both the total number of entities extracted by each system within individual report sections and the average number of entities extracted per report. To assess the breadth of information captured, we also counted the number of unique disease entities identified by each system across the dataset. Differences in the number of entities extracted per report between systems were statistically compared using paired t-tests across all model combinations, with Bonferroni correction applied to adjust for multiple comparisons.

For assertion detection, we report the distribution of extracted entities classified as *positive, negative, or uncertain* for each section of the CXR report, using the standardization criteria defined in Table 2. To evaluate whether the distribution of assertion types differed significantly between NLP systems, Chi-square tests of independence were performed separately for the Findings and Impression sections.

For CheXpert labels comparison, inter-model agreement for assertion classification was assessed using Fleiss' Kappa, calculated across all six NLP systems, four commercial systems (AWS, AZ, GC, and SP) and two open-source CXR-specific models (CheXbert and CheXpert), for each disease category. The assertion category *absent* was assigned when an NLP system did not detect a given label. Fleiss' Kappa values were computed under two conditions. In the first condition (*All*), all exams were included irrespective of whether the disease was detected or marked as *absent* by the models. In the second condition (*Excluding Absent*), exams were excluded if all six models predicted the disease as *absent*. This approach allows for a more accurate assessment of inter-model agreement in cases where the disease is actually present or asserted by at least one model, avoiding artificial inflation of agreement due to consistent non-detection.

To estimate model-specific assertion performance, a pseudo–ground truth was established using a majority voting strategy across outputs from all six NLP systems. For each disease entity, the consensus assertion label was determined by majority vote. If no majority was reached, the entity was assigned to the uncertain category. Assertion accuracy for each model was computed by comparing its predicted assertion category to the consensus label for each disease. A match was counted only when the assertion category exactly aligned with the consensus. Mean accuracy was reported separately for each disease, for each model, and as an overall mean, providing a summary of assertion performance.

**Results:**

**a) Named entity recognition by commercial clinical NLP models:**

*Table 3: Number of disease related entities extracted by each clinical NLP system for the Findings and Impressions sections of the CXR reports in the study dataset (n=95,008). Bolded numbers signify the highest count of extracted entities in the report sections.*

| Section | AWS | AZ | GC | SP |
|---|---|---|---|---|
| FINDINGS | 417,630 | **689,774** | 333,516 | 457,540 |
| IMPRESSION | 154,397 | **196,911** | 140,201 | 170,728 |
| All sections | 846,137 | **1,175,594** | 741,958 | 900,655 |

**Error! Reference source not found.** shows the total number of disease-related entities extracted by each system and their distribution for the Findings and Impressions sections of the CXR report, along with the overall counts including all sections (Clinical History, Comparison and Procedure Comments). Figure 2 shows the average number of disease related entities extracted per report by each clinical NLP system along with the standard deviation, whereas Figure 3 shows the unique number of disease related entities extracted by each system across the study dataset. All pairwise comparisons in the number of extracted entities per report were statistically significant (Bonferroni-adjusted p-value < 0.001). AZ extracted more entities overall (12.4), followed by SP (9.5), AWS (8.9), and GC (7.8), respectively. In terms of uniqueness, however, SP extracted considerably more entities (49,688) when compared to the other three systems. AZ and AWS extracted 31,543 and 27,216 unique entities, respectively, while GC had the lowest count with only 16,477 unique entities.

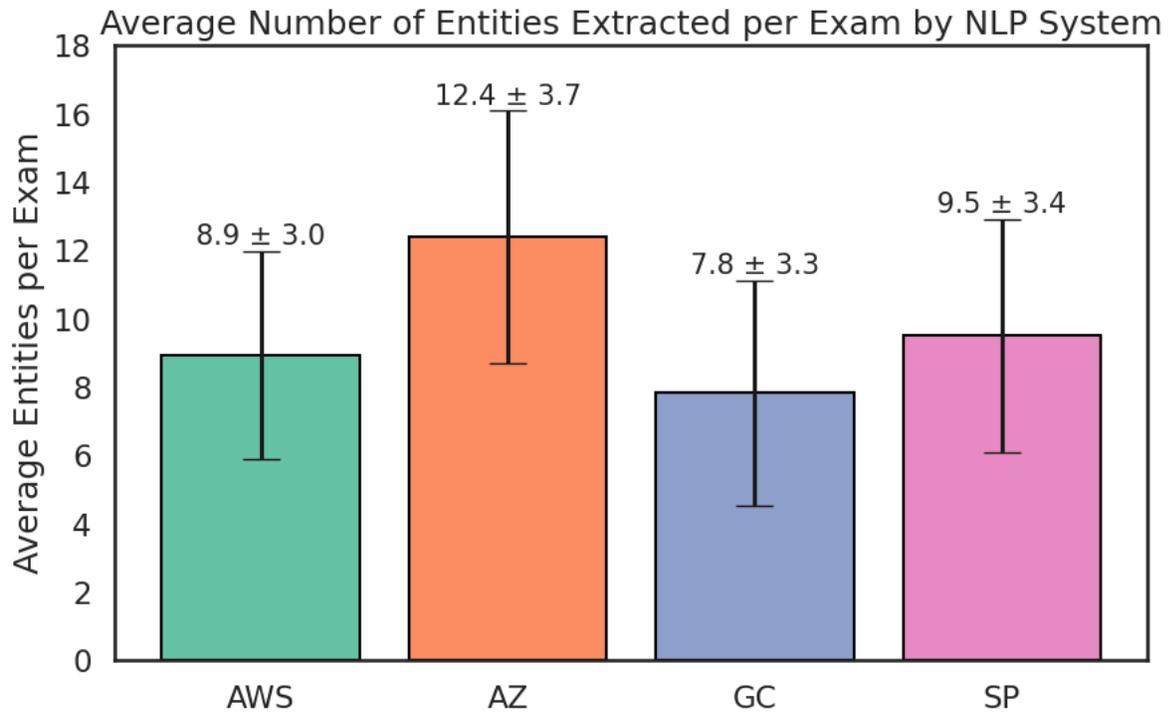

*Figure 2: The average number of disease related entities extracted per CXR report by each clinical NLP system. AWS – Amazon Medical Comprehend, AZ – Azure Text Analytics for Health, GC – Google Healthcare Natural Language API, SP – John Snow Labs' Spark NLP models for radiology. The whiskers represent the standard deviation from the average value.*

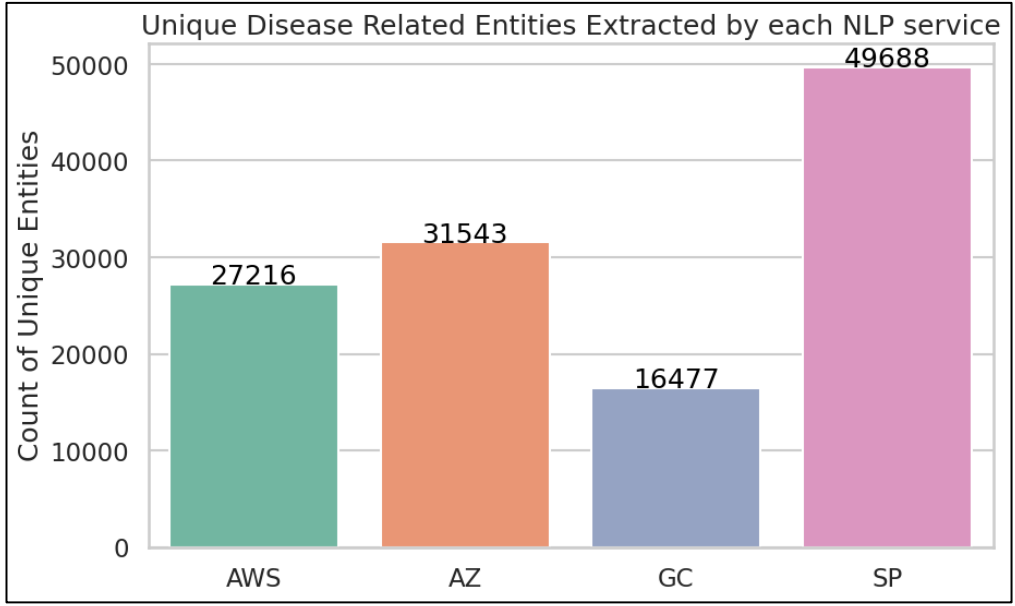

*Figure 3: Unique number of disease related entities extracted by each clinical NLP system on 95,008 pediatric CXR reports. AWS – Amazon Medical Comprehend, AZ – Azure Text Analytics for Health, GC – Google Healthcare Natural Language API, SP – John Snow Labs' Spark NLP models for radiology.*

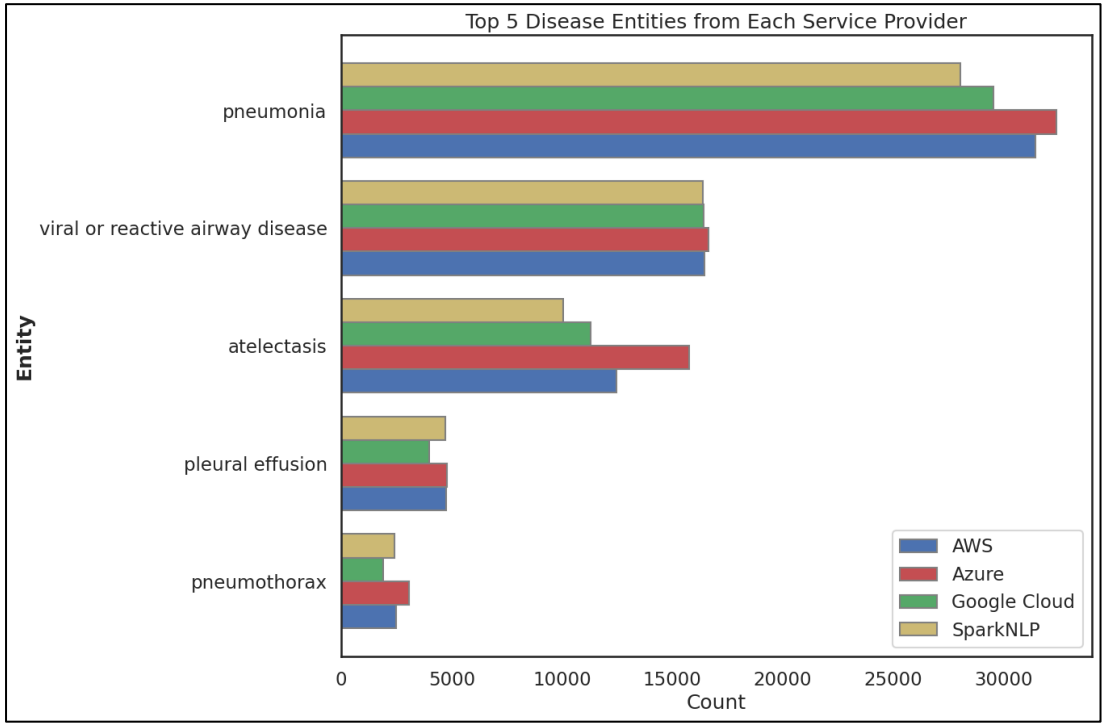

*Figure 4: Count of top 5 most frequently extracted disease entities by the 4 commercial NLP systems on the study dataset from the Impression section of the reports. AWS – Amazon Medical Comprehend, AZ – Azure Text Analytics for Health, GC – Google Healthcare Natural Language API, SP – John Snow Labs' Spark NLP models for radiology.*

Figure 4 illustrates the count of the top five most frequently detected diseases extracted by each of the four NLP systems from the Impression section of the reports - highlighting both detection frequency and inter-system variability.

*Pneumonia* is the most identified disease entity across all systems, with counts ranging from approximately 28,037 (SP) to 32,392 (AZ), reflecting a moderate spread of about 4,355 entities (≈14.4% variation). *Viral or reactive airway* disease shows the smallest variation, with counts tightly clustered between 16,460 (AWS) and 16,391 (SP) - just a 0.4% variation, indicating strong agreement. *Atelectasis*, however, shows the largest disparity, with AZ identifying nearly 15,774 instances, while SP reports only about 10,056, a spread of 5,718 entities (≈44.3% variation), suggesting significant disagreement. *Pleural effusion* has consistent counts between 3986 (GC) and 4793 (AZ), having a moderate spread with a difference of 807 (≈18.4% variation). *Pneumothorax* is the least commonly identified in the top 5, ranging between 1,885 (GC) and 3,065 (AZ) - a 1,180 difference (≈47.7% variation). These statistics suggest that while systems largely agree on certain entities, like *viral or reactive airway and pleural effusion*, others such as *atelectasis* and *pneumonia* show greater variability, likely due to differences in model sensitivity, training data, variability in medical term usage or recognition.

**b) Assertion detection by commercial clinical NLP models:**

*Table 4: Comparison of assertion detection performance across different models (AWS, AZ, GC, SP) for various sections in CXR reports. Percentages are rounded to one decimal place.*

| Assertion | System | FINDINGS Percentage (Assertion count / Total entities) | IMPRESSIONS Percentage (Assertion count / Total entities) |
|---|---|---|---|
| Positive | AWS | 51.2% (213,654 / 417,630) | 79.5% (122,652 / 154,397) |
| Positive | AZ | 72.3% (499,003 / 689,774) | 67.5% (132,857 / 196,911) |
| Positive | GC | 36.8% (122,789 / 333,516) | 61.0% (85,551 / 140,201) |
| Positive | SP | 70.5% (322,788 / 457,540) | 58.2% (99,342 / 170,728) |
| Negative | AWS | 48.3% (201,809 / 417,630) | 19.2% (29,617 / 154,397) |
| Negative | AZ | 26.1% (180,147 / 689,774) | 4.6% (9,125 / 196,911) |
| Negative | GC | 61.8% (206,052 / 333,516) | 21.8% (30,498 / 140,201) |
| Negative | SP | 26.9% (123,421 / 457,540) | 9.2% (15,777 / 170,728) |
| Uncertain | AWS | 0.5% (2,167 / 417,630) | 1.4% (2,128 / 154,397) |
| Uncertain | AZ | 1.5% (10,624 / 689,774) | 27.9% (54,929 / 196,911) |
| Uncertain | GC | 1.4% (4,675 / 333,516) | 17.2% (24,152 / 140,201) |
| Uncertain | SP | 2.5% (11,331 / 457,540) | 32.6% (55,609 / 170,728) |

Table 4 presents the distribution (Detected assertion entities / Total entities in report section) of assertion classifications for medical entities detected by the four models (AWS, AZ, GC, and SP) for the Findings and Impressions sections of chest X-ray (CXR) reports. In the Findings section, most entities were either *positive* or *negative* across all models, with *uncertain* classifications ranging from 0.5% (AWS) to 2.5% (SP). SP classified 70.5% of Findings as *positive,* whereas GC classified only 36.8%. For the Impression section, AWS classified 79.5% of entities as *positive*, while SP classified only 58.2%. *Negative* assertions were largest for GC (21.8%) and smallest for AZ (4.6%). *Uncertain* classifications were minimal for AWS (1.4%) but reached 32.6% for SP. The Chi-square test of independence showed that the distribution of entity assertions differed significantly between NLP systems in both the Findings and Impression sections ($p < 0.001$). These variations highlight differences in assertion detection performance, which may impact clinical decision support applications.

**c) Comparison with open-source CXR report labelers on CheXpert labels:**

Figure 5 shows the Fleiss' Kappa values for each of 12 disease categories and the *No Findings* category, based on the assertion statuses (positive, negative, uncertain and absent) assigned by all six NLP models. The highest agreement was observed for *pleural effusion* (Kappa: 0.89 in *All*; 0.59 in *Excluding Absent*), while the lowest was for e*nlarged cardiomediastinum* (Kappa: 0.25 in *All*; 0.05 in *Excluding Absent*). The mean Fleiss' Kappa across all categories were 0.68 ± 0.17 (*All*) and 0.35 ± 0.14 (*Excluding Absent*).

The individual assertion performance of the six NLP models against the consensus pseudo-ground truth calculated using majority voting for the CheXpert labels are shown in Table 5. The results are only based on the Impression section of the CXR reports. The lowest accuracy against the consensus was observed for the *Consolidation* category (14 ± 3) %, where AWS had the largest accuracy of 20% and GC had the smallest accuracy of 10%. Excluding No Findings, the highest mean accuracy was observed for Pleural Effusion (72 ± 6) %, with GC having the largest value of 82% and AWS having the smallest value of 62%. When considering the aggregate across all disease categories, SP out-performed other NLP systems with an accuracy of (76 ± 4) %, while AWS trailed with an accuracy of 50%. The mean overall accuracy was (62 ± 9) %. Both CheXbert and CheXpert had an overall accuracy of (56 ± 8) %.

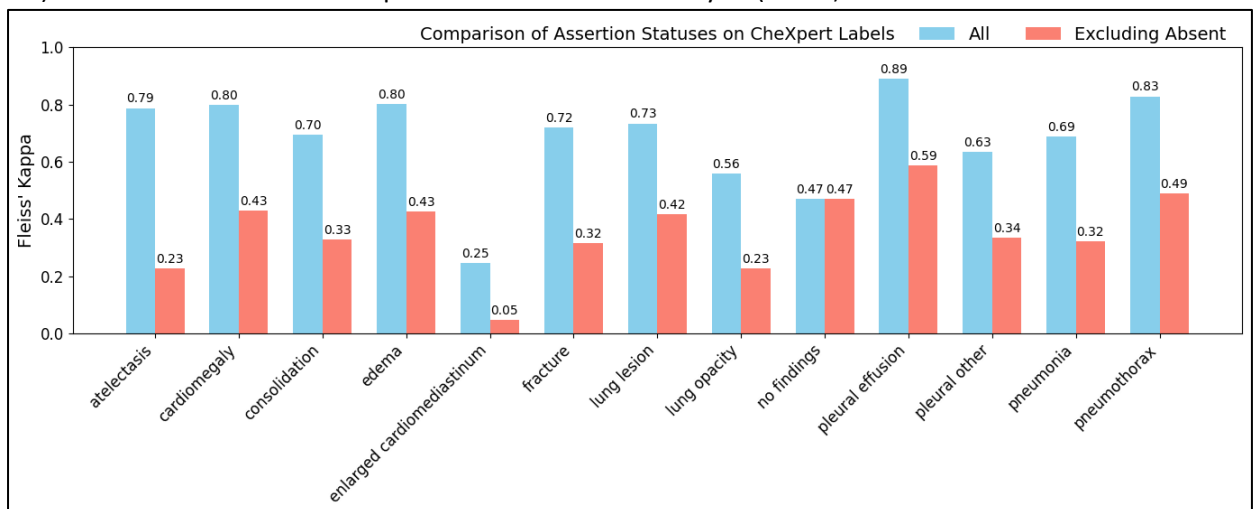

*Figure 5: Fleiss' Kappa values for each CheXpert label calculated across the six NLP systems under two conditions: (All) includes all exams regardless of whether the disease was detected; (Excluding Absent) excludes exams where the disease was not detected by any of the six NLP systems.*

*Table 5: Assertion accuracy (%) of the six NLP systems on the impression section of the study dataset with 95,008 chest X-ray (CXR) reports, evaluated against the consensus pseudo–ground truth on the CheXpert labels. Percentages are rounded to the nearest whole number.*

| Disease Category | AWS (%) | AZ (%) | GC (%) | SP (%) | CheXbert (%) | CheXpert (%) | Mean (%) |
|---|---|---|---|---|---|---|---|
| Atelectasis | 35 | 71 | 86 | 57 | 65 | 65 | 63 ± 15 |
| Cardiomegaly | 60 | 56 | 80 | 63 | 59 | 56 | 62 ± 8 |
| Consolidation | 20 | 12 | 10 | 13 | 12 | 17 | 14 ± 3 |
| Edema | 44 | 69 | 76 | 82 | 62 | 61 | 66 ± 12 |
| Enlarged Cardiomediastinum | 32 | 30 | 51 | 26 | 19 | 13 | 28 ± 12 |
| Fracture | 20 | 36 | 33 | 47 | 29 | 28 | 32 ± 8 |
| Lung Lesion | 47 | 65 | 61 | 70 | 55 | 55 | 59 ± 7 |
| Lung Opacity | 54 | 58 | 54 | 65 | 46 | 46 | 54 ± 7 |
| Pleural Effusion | 62 | 71 | 82 | 76 | 68 | 70 | 72 ± 6 |
| Pleural Other | 47 | 57 | 47 | 78 | 66 | 66 | 60 ± 11 |
| Pneumonia | 17 | 38 | 39 | 76 | 34 | 35 | 40 ± 18 |
| Pneumothorax | 40 | 52 | 56 | 43 | 42 | 41 | 46 ± 6 |
| No Findings | 72 | 98 | 70 | 89 | 67 | 67 | 77 ± 12 |
| Overall Mean | 50 ± 0 | 69 ± 5 | 63 ± 0 | 76 ± 4 | 56 ± 8 | 56 ± 8 | 62 ± 9 |

d) **Sample reports with discrepant assertion statuses extracted by the NLP systems for CheXpert labels:**

*Table 6: Variability in Assertion Among Commercial NLP Systems - Divergent NLP Interpretations of Radiology Reports Impressions.*

| Sample | Disease Label | Report Impression | Positive | Uncertain | Negative |
|---|---|---|---|---|---|
| 1 | pneumothorax | 1. Pectus bars in place.  2. No appreciable pneumothorax on the left. | SP | CheXpert, CheXbert, AWS, GC | AZ |
| 2 | pneumonia | Findings consistent with viral or reactive airways disease without focal pneumonia. | AZ | SP | CheXpert, CheXbert, AWS, GC |
| 3 | cardiomegaly | No acute cardiopulmonary abnormality with stable cardiomegaly and fracture of one of the pacemakers leads. | CheXpert, CheXbert, AZ, GC | AWS | SP |
| 4 | consolidation | Right suprahilar opacity may represent developing consolidation, superimposed on findings of viral or reactive airways disease. | AWS, AZ, SP | CheXpert, CheXbert, GC | |
| 5 | atelectasis | Viral/reactive airway disease, superimposed with right upper and left lower lobe airspace | AWS, AZ, GC | CheXpert, CheXbert | SP |

| | | disease such as atelectasis/pneumonia. | | | |
|---|---|---|---|---|---|
| 6 | edema | Mildly prominent pulmonary vasculature. Cardiac size appears mildly enlarged. No focal airspace disease or overt pulmonary edema is suspected. | GC | AZ | CheXpert, CheXbert, AWS, SP |
| 7 | lung lesion | No consolidation. Small oval lucency in the left upper lobe. It is not clear whether this represents superimposed shadows (mock effect) or a true finding. A short-term follow-up two-view chest x-ray is suggested to evaluate the persistence of this finding. If it does persist, it may represent a tiny bleb, bulla, or pneumatocele. The surrounding lung appears normal making a cavitary lesion less likely. | CheXpert, AWS, GC | CheXbert, AZ, SP | |
| 8 | lung opacity | Poorly defined left lower lobe opacity concerning for developing pneumonia. | CheXpert, CheXbert, AWS | AZ, GC, SP | |

Table 6 describes variability in assertion among commercial NLP systems. It illustrates how clinical NLP systems interpret assertions based on linguistic features in the impression section. In case 1 for *pneumothorax*, the impression "No appreciable pneumothorax on the left" led to mixed interpretations, with SP classifying it as positive while most others marked it negative. For *pneumonia* in case 2, the phrase "without focal pneumonia" resulted in most systems classifying it as negative, with only AZ marking it positive and SP classifying it as uncertain. The *cardiomegaly* example (case 3) with " No acute cardiopulmonary abnormality with stable cardiomegaly" term led four systems to classify it as positive, while AWS remained uncertain, and SP classified it as negative. In case 4, the hedging phrase "may represent developing consolidation" resulted in three systems classifying *consolidation* as uncertain and three as negative. Similarly, in case 5 for *atelectasis*, the complex description of "airspace disease such as atelectasis/pneumonia" divided systems, with three positive, two uncertain, and one negative. Notably, in case 8, the phrase "opacity concerning for developing pneumonia" led to a split between positive and uncertain classifications of *lung opacity*. These findings demonstrate how radiological language, especially uncertainty markers, qualifiers, and alternative explanations consistently challenge commercial NLP systems, with each interpreting probabilistic medical language and hedging expressions differently.

**Discussion:**

This study is one of the first to quantify and compare the performance of commercial clinical NLP systems on a large independent dataset, and the first to do so on a study sample composed of pediatric chest radiograph reports. Commercial clinical NLP systems from three major cloud providers, namely Amazon, Google, and Azure, along with a radiology specific model from a well-known vendor, John Snow Labs, were analyzed on the tasks of named entity recognition and assertion detection. A standardization algorithm was developed to map the entities extracted by these systems to disease labels defined by the

CheXpert framework. The CheXpert and CheXbert models provided a set of disease categories that served as a reference for further comparison. Model outputs were evaluated using a consensus ground truth derived from the outputs of all six NLP systems using a majority voting approach.

a) **Variability in entity extraction**

Comparison of the disease related entities extracted by the four commercial NLP systems revealed considerable variability in the mean number of entities per report and the number of unique entities detected. AZ recorded the highest mean number of entities per report, while SP produced the greatest number of unique entities. In addition, the top five most frequent disease or diagnosis entities reported had variable agreement between the systems. Viral or reactive airway disease had near perfect agreement (< 1% difference in counts) between the four systems, while pneumothorax and atelectasis had large differences (> 40% difference in counts). These results illustrate the inconsistencies in named entity recognition and underline the importance of applying standardization techniques, such as the use of regular expressions, to achieve meaningful comparisons.

b) **Variability in assertion detection**

Large variability was also observed in assertion detection performance by the four NLP systems. Firstly, each NLP system reported assertion status in its own way, and the definitions were not consistent. Secondly, once the assertion statuses were grouped into positive, negative, and uncertain categories, significant differences in their distributions were observed between the systems for both Findings and Impression sections. Especially, for the Findings section, the NLP systems reported 0.5% (AWS) to 2.5% (SP) of the entities as *uncertain,* whereas for the Impression section, the number of uncertain entities varied between 1.4% (AWS) to 32.6% (SP).

c) **Performance on CheXpert labels**

When evaluating disease labels based on the CheXpert framework, inter-model agreement was substantial when *absent* predictions were included (mean Kappa = 0.68) but dropped to fair agreement when those cases were excluded (mean Kappa = 0.35), highlighting variability in assertion classification when disease presence was detected by at least one model. The overall mean accuracy across all six models was 62% with a standard deviation of 9%. Mean assertion accuracy for individual diseases ranged from 14% for *consolidation* to 77% for *pleural effusion*. Poor performance for consolidation may stem from the wide variety of expressions used in reports. As a descriptive imaging finding, it is mentioned in more variable and nuanced ways and can appear in multiple conditions such as pneumonia or pulmonary edema. Higher accuracy for pleural effusion possibly results from its more explicit mentions.

Among the disease categories, SP outperformed the other systems with an accuracy of 76%. CheXpert and CheXbert performed similarly for most categories, with notable differences only for *consolidation* and *enlarged cardiomediastinum*. Assertion accuracy varied by category, with SP performing best in six categories, GC in five, and AWS and AZ each in one category.

d) **Limitations and future directions:**

Investigation of cases with discrepant assertion statuses revealed that linguistic nuances influence each NLP model differently. No single system demonstrated uniform superiority. Instead, performance varied by disease category and the contextual phrasing within report sections. These findings highlight the potential benefits of model ensembling, where the complementary strengths of different models can be leveraged to improve overall robustness and accuracy. This is reinforced by the variability observed across cases in Table 6. Each system shows strengths and limitations depending on the linguistic structure, diagnostic ambiguity, and type of condition described. For instance, AZ accurately identified the absence of *pneumothorax* in sample 1 but failed to detect the absence of *pneumonia* in sample 2. AWS correctly recognized the absence of *pneumonia* in sample 2 but was uncertain about *cardiomegaly* in sample 3 due to ambiguous language, although clinically this indicates a positive finding. SP successfully detected the absence of edema in sample 6 but missed the absence of *pneumothorax* in sample 1. Meanwhile, CheXpert, CheXbert, and GC all correctly flagged uncertain consolidation in sample 4 but failed to identify the absence of *pneumothorax* in sample 1. This inconsistency highlights that NLP performance is highly context-dependent, shaped by how each system interprets medical uncertainty, negation, and diagnostic probability.

This observation underscores the necessity for thorough validation of clinical NLP systems for specific use cases and institutions, as well as careful evaluation of non-standard definitions for uncertainty before these systems are deployed. One limitation of this study is that the AWS did not provide a categorical assertion status and offered only a separate negation attribute. Confidence values were used to assign uncertainty in these cases. The latest version of Amazon's Comprehend Medical NLP system includes an attribute for low confidence that may provide a better estimation of uncertainty. In addition, the large dataset precluded manual annotation of CheXpert labels. Therefore, a pseudo ground truth derived from all six NLP systems was used. Cases without a clear majority were assigned to the uncertain category. Although the pseudo ground truth is useful for comparing systems against each other, performance metrics may differ when compared to labels assigned by radiologists.

Future research should incorporate manual annotations and investigate the effect of these variabilities on downstream clinical and research applications to ensure that inherent errors do not propagate and adversely affect patient care. Additionally, this study is limited to only pediatric chest X-ray reports, without including adult cases or a broader range of imaging modalities. While this limits the direct generalizability of the findings, the insights gained are likely applicable across other modalities and populations.

**Conclusion:**

Significant variability exists in named entity recognition and assertion assessment on CXR radiology reports across different NLP systems. These differences arise from variations in the clinical concept definitions inherent to each system and the absence of a uniform approach to quantifying uncertainty expressions. In addition, the diversity in how CXR reports describe radiological findings and impressions results in systems performing optimally under different conditions. Consequently, the use of automated NLP systems for labeling imaging exams and for downstream applications such as outcome predictions must be accompanied by careful task specific evaluation and oversight.